# Counting with Confidence: Accurate Pest Monitoring in Water Traps

Xumin Gao[1]*, Mark Stevens[2], Grzegorz Cielniak[1]

*Lincoln Centre for Autonomous Systems, University of Lincoln, Lincoln, UK*
*British Beet Research Organisation, Colney Lane, Norwich, UK*
*25766099@students.lincoln.ac.uk*

**Abstract**: Accurate pest population monitoring and tracking their dynamic changes are crucial for precision agriculture decision-making. A common limitation in existing vision-based automatic pest counting research is that models are typically evaluated on datasets with ground truth but deployed in real-world scenarios without assessing the reliability of counting results due to the lack of ground truth. To this end, this paper proposed a method for comprehensively evaluating pest counting confidence in the image, based on information related to counting results and external environmental conditions. First, a pest detection network is used for pest detection and counting, extracting counting result-related information. Then, the pest images undergo image quality assessment, image complexity assessment, and pest distribution uniformity assessment. And the changes in image clarity caused by stirring during image acquisition are quantified by calculating the average gradient magnitude. Notably, we designed a hypothesis-driven multi-factor sensitivity analysis method to select the optimal image quality assessment and image complexity assessment methods. And we proposed an adaptive DBSCAN clustering algorithm for pest distribution uniformity assessment. Finally, the obtained information related to counting results and external environmental conditions is input into a regression model for prediction, resulting in the final pest counting confidence. To the best of our knowledge, this is the first study dedicated to comprehensively evaluating counting confidence in counting tasks, and quantifying the relationship between influencing factors and counting confidence through a model. Experimental results show our method reduces MSE by 31.7% and improves $R^2$ by 15.2% on the pest counting confidence test set, compared to the baseline built primarily on information related to counting results.

## 1. INTRODUCTION

Pests harm crops and reduce yields. Current control relies on regular pest counting to monitor population trends and guide measures like insecticide spraying. Therefore, accurately and promptly counting the number of pests is crucial. While there has been considerable research on automatic pest counting [1], these studies often lack a measure of confidence due to the absence of ground truth in actual tests, making it difficult to assess the reliability of the counting results. In our previous research [2], we proposed a novel aphid counting method using interactive stirring actions and a counting confidence evaluation system to overcome undercounting due to occlusion. However, it only considered the mean detection confidence of aphid bounding boxes, the predicted number of aphids, and the image average gradient magnitude as factors influencing counting confidence. These factors are mainly based on information related to counting results and do not take into account external environmental factors, such as lighting changes, background complexity, and others. Through literature review, we identified key external factors affecting counting confidence: image quality, image complexity, and object distribution uniformity. Generally, higher image quality enhances target identification, leading to more accurate counts. Increased image complexity, characterized by denser objects and greater background noise, complicates target recognition, causing incorrect and missed counts. Uneven object distribution increases the likelihood of occlusion, leading to incorrect and missed counts. Additionally, while some existing research explores how various experimental conditions affect counting performance [3-8], these studies have two main limitations: 1) They only examine monotonic trends in counting performance under different factors, without establishing a quantitative model that can accurately reflect the relationship between influencing factors and counting performance. 2) They analyze individual factors separately, without considering their combined impact on counting performance. To this end, in this paper, we propose a method for comprehensively evaluating counting confidence, based on information related to counting results and external environmental conditions, to assess the reliability of pest counting results in water traps. Specifically, 1) We utilized the detection network from our previous work [2] for pest detection and counting. 2) We designed a hypothesis-driven multi-factor sensitivity analysis to identify the optimal image quality assessment (IQA) and image complexity assessment (ICA) methods, and applied them to evaluate image quality and complexity. 3) We proposed an adaptive DBSCAN clustering algorithm for pest distribution uniformity (PDU) assessment to mitigate over-clustering and under-clustering issues inherent in the standard DBSCAN algorithm. 4) To quantify changes in image clarity caused by stirring during image acquisition, we calculated the average gradient magnitude (AGM), consistent with our previous work [2]. 5) For pest counting confidence evaluation, we developed a regression model that evaluates the counting confidence based on information related to counting results and external environmental conditions.

## 2. RELATED WORK

*2.1 Impact of counting result-related information on object counting*

Counting result-related information based on the detection method includes the confidence of bounding boxes and the overall predicted count. Low confidence in a predicted bounding box indicates unreliable detections, increasing the likelihood of incorrect counts. A higher overall predicted count signifies greater object density, which raises the probability of occlusion and overlap, leading to incorrect and missed counts. Our previous work [2] also confirmed these findings. Therefore, counting result-related information, being the most direct indicators, can serve as fundamental factors for evaluating counting confidence.

*2.2 The impact of image quality on object counting*

Image quality has a significant impact on object counting performance. Low-quality images (such as blurriness, noise) can affect the accuracy of feature extraction, especially due to the loss of details, which leads to incorrect counts and missed counts, thus making the counting results unreliable. Li et al. [3] used a detection network combined with a tracking algorithm for counting passengers disembarking from buses. Their experimental results showed that, compared to environments with bright daylight, the system's ability to detect and track passengers in dark environments significantly declined, making the counting results unreliable. Yang et al. [4] proposed a detection network combined with a multi-object tracking algorithm for counting fish fry. Their results showed that as the fish fry's speed increased from 0.24 m/s to 0.84 m/s, counting accuracy dropped from 97.88% to 87.89%, due to image blurring.

*2.3 The impact of image complexity on object counting*

Image complexity is one of the key factors influencing object counting performance. In highly complex scenarios, dense object distributions and substantial background noise often lead to severe occlusion, significantly increasing the difficulty of counting. Arteta et al. [5] explored challenges in counting penguins in the wild, finding that occlusion, cluttered backgrounds negatively impacted counting accuracy. Their results showed that counting errors increased with the number of penguins, as occluded penguins were often missed. Background rocks also caused confusion, leading to incorrect counts. Shao et al. [6] explored factors contributing to inaccurate crowd counting in outdoor scenarios, including mutual occlusion between individuals and diverse scenario distributions, which hinder model recognition and generalization. To address these issues, they used synthetic data to simulate different occlusion levels and crowd densities, improving data availability and enhancing the model's performance in real-world scenarios.

*2.4 The impact of object distribution uniformity on object counting*

In real-world object counting tasks, object distribution uniformity is a critical factor influencing counting performance. The closer the objects are to each other, the more severe the occlusion between them, which exacerbates incorrect counts and missed counts in counting. Ioannis, et al. [7] developed an insect counting system in E-Traps. Experimental results showed that as insect density increased, the MAE (Mean Absolute Error) of models rose, indicating reduced counting accuracy in high-density scenarios. GAO et al. [8] proposed a hybrid network combining detection and density map estimation for aphid counting. Experimental results showed the model achieved a MAE of 2.93 and RMSE (Root Mean Squared Error) of 4.01 for standard-density aphids, but these values increased dramatically to 34.19 (MAE) and 38.66 (RMSE) in high-density scenarios.

## 3. METHOD

*3.1 Overview of our proposed method*

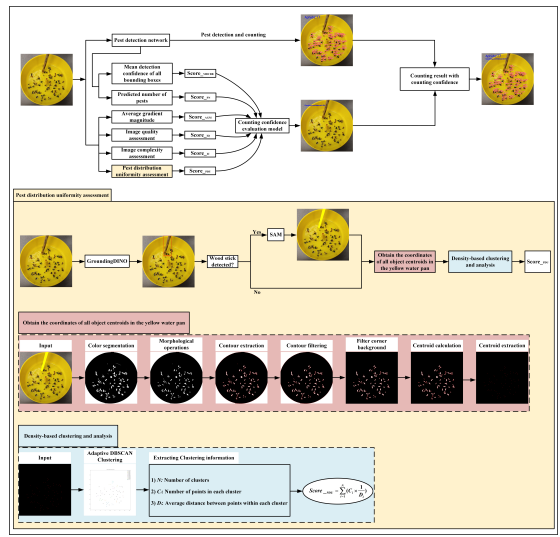

Fig. 1. The proposed counting confidence evaluation method

As shown in Fig. 1, we first input a pest image captured from a yellow water trap under stirring conditions into a pest detection network to identify pests and obtain the counting result. This process allows us to extract counting result-related information, including the mean detection confidence of bounding boxes (MDCBB) and the predicted number (PN) of pests, represented as $Score\_{MDCBB}$ and $Score\_{PN}$. We also assess image quality, image complexity, and pest distribution uniformity, resulting in scores $Score\_{IQ}$, $Score\_{IC}$ and $Score\_{PDU}$. Additionally, consistent with our previous work [2], we quantify changes in image clarity induced by stirring through the calculation of the average gradient magnitude, resulting in $Score\_{AGM}$. We then input all these scores into a counting confidence evaluation model that uses regression to assess the confidence of the pest count results. Finally, we obtain the counting result along with the associated counting confidence. The following sections detail the pest detection network (Section 3.2), the selection of optimal IQA and ICA methods (Section 3.3), the proposed pest distribution uniformity assessment method (Section 3.4), and the designed counting confidence evaluation model (Section 3.5).

*3.2 Pest detection network*

For the pest detection network, we use an improved Yolov5 from our previous aphid counting work [2], with the key difference being the absence of the split-merge strategy in this paper. This is because we employ bionic insects to simulate real pests in this paper, which are larger in size (0.75 cm to 2 cm) compared to real-world pests. Using a split-merge strategy would fragment many complete pests across image sub-blocks, compromising detection and counting accuracy. This rationale will be validated in the experiments presented in Section 4.

*3.3 Image quality assessment and image complexity assessment*

We design a hypothesis-driven multi-factor sensitivity analysis method to select the optimal IQA and ICA methods. The design approach is outlined as follows: 1) Establishing hypotheses. Our hypotheses are based on four key factors that may affect image quality and image complexity during the pest counting process. These factors include: temporal changes, stirring speed, the presence of soil, and pest density. The specific hypotheses are as follows: a) Temporal changes: Image quality decreases during the stirring phase compared to the static phase; image complexity increases during the stirring phase compared to the static phase. b) Stirring speed: Image quality decreases with an increase in stirring speed; image complexity increases with an increase in stirring speed. c) Presence of soil: Image quality decreases when soil is present compared to when soil is absent; image complexity increases when soil is present compared to when soil is absent. d) Pest density: Image quality decreases with an increase in pest density; image complexity increases with an increase in pest density. 2) Verifying hypotheses. We first collect dataset under different conditions: temporal stages (static and stirring), stirring speeds (low, medium, high), presence or absence of soil, and varying pest densities (low and high). Next, we use multiple mainstream IQA methods (e.g., HYPERIQA, DBCNN, MUSIQ, CLIP-IQA, LIQE, NIQE, PIQE, BRISQUE) [9] and ICA methods (e.g., Entropy, CNN-based methods, Edge density) [10] to assess the image quality and complexity. We then analyze the differences in the mean scores of image quality and complexity across the conditions mentioned above to evaluate the sensitivity of various IQA and ICA methods. Additionally, independent sample T-tests (for temporal changes, soil presence/absence, and pest density) and one-way ANOVA (for stirring speed) are conducted to validate our hypotheses. Finally, we compare the sensitivity and statistical test results of different IQA and ICA methods under varying conditions to select the optimal IQA and ICA methods, which are then used to assess image quality and complexity, yielding image quality scores *Score_IQ* and image complexity scores *Score_IC*.

*3.4 Pest distribution uniformity assessment*

In this paper, we propose an adaptive DBSCAN clustering algorithm for pest distribution uniformity assessment. As shown in the pest distribution uniformity assessment section of Fig. 1, we first use GroundingDINO [11] with the prompt "wood stick" to detect the stirring tool. If detected, SAM (Segment Anything Model) [12] segments it and adjusts its color to match the yellow trap background for easier removal in the next color segmentation step, reducing background noise. If no wood stick is detected, the image remains unprocessed. Next, we apply image processing techniques, including color segmentation, morphological operations, contour filtering, background corner filtering, and centroid calculation, centroid extraction, to obtain the centroids of all objects in the yellow water trap. Next, we apply an adaptive DBSCAN clustering algorithm to cluster object centroids. It is important to emphasize that the adaptability of the proposed adaptive DBSCAN clustering lies in its radius parameter, which is determined by the average size of detected pests in the image. Specifically, after obtaining detection results, we compute the average size of all predicted bounding boxes and use it to set the clustering radius. Finally, we extract the clustering information from the clustering results, including: 1) The number of clusters $N$. 2) The number of points in each cluster $C_i$. 3) The average distance between points within each cluster $D_i$. And we use Eq. (1) to compute the pest clustering score, which represents the pest distribution uniformity score *Score_PDU*.

$$Score_{-PDU} = \sum_{i=1}^{N}(C_i \times \frac{1}{D_i}) \quad (1)$$

*3.5 Counting confidence evaluation model*

We develop a pest counting confidence evaluation model that can comprehensively assess pest counting confidence, based on influencing factors, including the mean detection confidence of all bounding boxes, the predicted number of pests, average gradient magnitude, image quality, image complexity, and pest distribution uniformity. Specifically, we analyze scatter plots to determine whether these factors have linear or nonlinear relationships with counting confidence. Based on this, we choose either a linear or nonlinear model to develop the pest counting confidence evaluation model. It should be specifically noted that during the development of the pest counting confidence evaluation model, we calculate counting confidence for each training image by comparing the pest detection network's predictions with ground truth labels. The counting confidence is evaluated using the Jaccard index, defined as *TP/(TP+FP+FN)*, where *TP*, *FP*, and *FN* denote the number of true positives, false positives, and false negatives, respectively.

## 4. EXPERIMENTS AND RESULTS

*4.1 Dataset*

Due to pests' short seasonal lifespan and weather sensitivity, collecting large datasets is challenging. To address this, we simulated real-world pest counting scenarios in a yellow water trap. We placed various bionic insects and soil into the trap, designating one insect type as the target pest while others and soil acted as interference. The trap was stirred with a wood stick, and images were captured every two seconds using a smartphone, which was mounted on a support stand during the data collection process. A complete data collection process is as follows: An image is captured at $T_0$ (0s) before stirring begins. At $T_1$ (2s), the wood stick is introduced into the yellow water trap, and stirring continues until $T_2$, when it stops. At $T_3$ ($T_2 + 1$), the stirring tool is removed, and image capture continues until $T_4$, when the water surface becomes nearly calm, yielding a complete image sequence.

Table 2. The results of hypothesis testing for IQA

| | Metric | HYPERIQA | DBCNN | MUSIQ | CLIPIQ | LIQE | NIQE | PIQE | BRISQUE |
|---|---|---|---|---|---|---|---|---|---|
| Temporal changes | Mean_Diff | 0.1739 | 0.2265 | -0.1143 | 0.1039 | -0.1449 | **-0.1581** | -0.0002 | -0.1427 |
| | P_value | 0.0027 | 0.0002 | 0.0249 | 0.0774 | 0.0137 | 0.0059 | 0.9968 | 0.0176 |
| Stirring speed | Mean_Diff (Med_Low) | -0.1607 | -0.1719 | -0.0555 | -0.1843 | -0.1263 | 0.0132 | 0.0440 | -0.0380 |
| | Mean_Diff (High_Low) | -0.1389 | -0.2600 | -0.1396 | -0.2877 | -0.3076 | 0.0599 | 0.1099 | -0.1549 |
| | Mean_Diff (High_Med) | 0.0219 | -0.0881 | -0.0840 | -0.1034 | -0.1813 | 0.0467 | 0.0659 | -0.1170 |
| | P_value | 0.0220 | 0.0001 | 0.0096 | 0.0000 | 0.0000 | 0.6385 | 0.2419 | 0.0769 |
| Presence of soil | Mean_Diff | -0.1112 | 0.1006 | -0.4647 | -0.5496 | -0.3448 | **-0.6829** | -0.1776 | -0.5611 |
| | P_value | 0.0925 | 0.0555 | 0.0000 | 0.0000 | 0.0000 | 0.0068 | 0.0000 | 0.0000 |
| Pest density | Mean_Diff | 0.1576 | 0.1411 | -0.0372 | **-0.1785** | 0.1275 | -0.0905 | -0.0279 | -0.1737 |
| | P_value | 0.0000 | 0.0000 | 0.0581 | 0.0000 | 0.0000 | 0.0000 | 0.1976 | 0.0000 |

To diversify the data, we varied experimental conditions, including pest quantities, stirring speeds, and soil presence. Notably, the decision to stop stirring and stop capturing images is based on human visual perception, causing T2, T3, and T4 to vary across different data groups.

We collected two datasets for pest detection and counting confidence evaluation, respectively. For pest detection, we collected 21 sets of data (410 images) under varying conditions, and split them into training, validation, and test sets (8:1:1). For counting confidence evaluation, we collected 35 sets of data (890 images) under varying conditions, and split them into training and test sets (7:3). Among them, 13 sets were used to select optimal IQA and ICA methods for testing the hypotheses in Section 3.3, designed using single-variable control with variations in pest densities (10, 20, 30, 40, 50, 60, 70, 80), stirring speeds (low, medium, high), and soil presence. To ensure fairness, we strictly controlled the timings during the data collection of these 13 sets, maintaining consistency in T2, T3, and T4 throughout the entire process.

*4.2 Implementation details*

All experiments were conducted using Python 3.8.13 and PyTorch 1.12.1. The pest detection model was trained with the same parameters as our previous work [2].

*4.3 Evaluation of pest detection network*

We conducted pest detection tests on the test set of the pest detection dataset. The evaluation metrics included AP@0.5, the number of TP, FP, FN, and the mean counting confidence (MCC). The results are shown in Table 1.

Table 1. The comparison results of detecting pests using different networks on the test set of pest detection dataset

| Method | AP@0.5 (%) | TP | FP | FN | MCC (%) |
|---|---|---|---|---|---|
| Yolov5 | 97.4 | 556 | 61 | 22 | 87 |
| [2] | 97 | 554 | 124 | 24 | 79 |
| Ours | 97.1 | 549 | 15 | 29 | **92.6** |

From Table 1, all detection networks achieve similar AP@0.5 (~97%) on the test set. However, our network achieves the highest MCC (92.6%), surpassing Yolov5 by 5.6% and [2] by 13.6%. While TP and FN are comparable across models, our network produces the fewest FP (15 vs. 61 for Yolov5 and 124 for [2]).

*4.4 Selection of the optimal IQA and IC methods*

As described in Section 4.1, we used 13 sets of data to validate the hypotheses in Section 3.3. For temporal stages, we selected the image at 0s for the static phase and images from 2s to T$_2$ for the stirring phase. For pest densities, datasets with densities ≤ 40 were grouped as low-density, and those > 40 as high-density. The results of hypothesis testing for IQA and ICA are shown in Table 2 and Table 3.

Table 3. The results of hypothesis testing for ICA

| | Metric | Edge density | Entropy | CNN |
|---|---|---|---|---|
| Temporal changes | Mean_Diff | -0.0163 | **0.2204** | 0.1734 |
| | P_value | 0.8024 | 0.0007 | 0.0071 |
| Stirring speed | Mean_Diff (Med_Low) | -0.0780 | 0.0911 | 0.0055 |
| | Mean_Diff (High_Low) | -0.2388 | 0.0096 | 0.0071 |
| | Mean_Diff (High_Med) | -0.1608 | -0.0816 | 0.0016 |
| | P_value | 0.0000 | 0.4202 | 0.9935 |
| Presence of soil | Mean_Diff | **0.8294** | 0.7521 | 0.5135 |
| | P_value | 0.0000 | 0.0000 | 0.0000 |
| Pest density | Mean_Diff | **0.3741** | 0.3265 | 0.3442 |
| | P_value | 0.0000 | 0.0000 | 0.0000 |

From Table 2, it can be observed that 1) Image quality scores for MUSIQ, LIQE, NIQE, and BRISQUE are significantly lower during stirring than in the static phase (Mean_Diff < 0, p_value < 0.05), supporting the hypothesis that "Image quality decreases during the stirring phase compared to the static phase". NIQE shows the largest difference (0.1581), followed by LIQE (0.1449), BRISQUE (0.1427), and MUSIQ (0.1143). In contrast, HYPERIQA and DBCNN show the opposite trend, while PIQE and CLIPIQA are insensitive to stirring (p_value > 0.05). 2) DBCNN, MUSIQ, CLIPIQA, and LIQE respond to stirring speed, showing higher image quality scores at high speeds. In contrast, HYPERIQA, NIQE, PIQE, and BRISQUE are insensitive to varying speeds (p_value > 0.05). Therefore, the hypothesis that "Image quality decreases with increasing stirring speed" is not supported. 3) Image quality scores for MUSIQ, CLIPIQA, LIQE, NIQE, PIQE, and BRISQUE significantly decrease in the presence of soil (Mean_Diff < 0, p_value < 0.05), supporting the hypothesis that "Image quality decreases in the presence of soil compared to the absence of soil". NIQE shows the largest drop (0.6829), followed by BRISQUE (0.5611), CLIPIQA (0.5496), MUSIQ (0.4647), LIQE (0.3448), and PIQE (0.1776). HYPERIQA and DBCNN are insensitive to soil conditions (p_value > 0.05). 4) The effect of pest density on image quality varies. CLIPIQA, BRISQUE, and NIQE have higher scores for low-density images, with differences of 0.1785, 0.1737, and 0.0905, respectively. In contrast, HYPERIQA, DBCNN, and LIQE score higher for high-density images. MUSIQ and PIQE are insensitive to changes in pest density (p_value > 0.05). Therefore, the hypothesis that "Image quality decreases with increasing pest density" is only partially supported. Overall, NIQE emerges as the optimal IQA method due to its significant sensitivity and consistency across multiple hypotheses. It shows the largest differences in image quality between the stirring and static phases (0.1581) and between the presence and absence of soil (0.6829). Additionally, NIQE also supports the hypothesis that "Image quality decreases with increasing pest density".

From Table 3, it can be observed that 1) The complexity assessment scores for Entropy and CNN methods significantly increase during the stirring phase, supporting the hypothesis that "Image complexity increases during the stirring phase compared to the static phase" (Mean_Diff > 0, p_value < 0.05). Specifically, the mean difference for Entropy is 0.2204, and for CNN, it is 0.1734. However, the complexity of Edge density slightly decreases during the stirring phase (Mean_Diff < 0) and does not support the hypothesis (p_value > 0.05). 2) The hypothesis that "Image complexity increases with increasing stirring speed" is not supported. Specifically, the complexity of Edge density is significantly lower at high stirring speeds compared to low speeds, contradicting the hypothesis. Additionally, Entropy and CNN show no significant differences between the two speed groups (p_value > 0.05), further failing to support the hypothesis. 3) All methods show a significant increase in complexity in the presence of soil (Mean_Diff > 0, p_value < 0.05). Specifically, the mean differences in complexity scores are 0.8294 for Edge density, 0.7521 for Entropy, and 0.5135 for CNN. Therefore, the hypothesis that "Image complexity increases in the presence of soil compared to the absence of soil" is supported. 4) All methods show a significant increase in complexity under high pest density conditions (Mean_Diff > 0). Specifically, the differences are 0.3741 for Edge density, 0.3442 for CNN, and 0.3265 for Entropy. Thus, all methods support the hypothesis that "Image complexity increases with increasing pest density" (p_value < 0.05). Considering the support and sensitivity across the four hypotheses, Entropy proves to be the optimal method for ICA. It shows significant sensitivity and consistency, with the highest mean difference (0.2204) in the hypothesis "image complexity increases during the stirring phase compared to the static phase" and ranking second in "image complexity increases in the presence of soil compared to the absence of soil" (0.7521) and "image complexity increases with increasing pest density" (0.3265). While Edge density ranks first in two hypotheses, it does not support the well-founded hypothesis that "Image complexity increases during the stirring phase compared to the static phase" revealing certain limitations.

### 4.6 Evaluation of pest conting confidence model

**Model development.** We used the pest detection network to detect and count pests in each image from the training set, extracting the corresponding scores for the mean detection confidence of all bounding boxes and the predicted number. Simultaneously, we applied the optimal IQA method (NIQE) and ICA method (Entropy), along with the proposed pest distribution uniformity assessment method, to calculate image quality, complexity, and pest distribution uniformity scores for each image. Additionally, the average gradient magnitude was calculated for each image. The scatter diagrams of the scores of each influencing factor against counting confidence are plotted, as shown in Fig. 2.

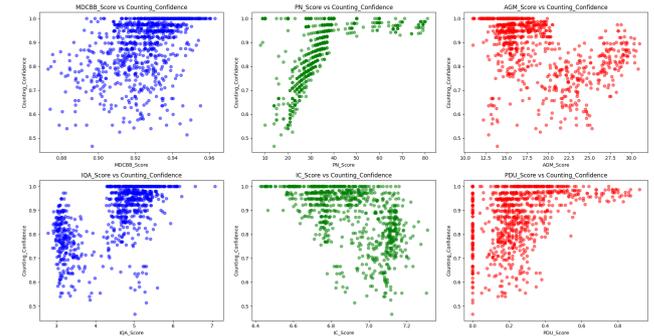

Fig. 2. The scatter diagrams of each influencing factor's scores against counting confidence

From Fig. 2, it can be observed that the relationships between influencing factors and the counting confidence all exhibit a curved trend, indicating a nonlinear characteristic. This indicates a nonlinear relationship between these influencing factors and the counting confidence. Therefore, we selected a nonlinear model, polynomial regression, to train the pest counting confidence model.

**Model evaluation.** We evaluated our proposed pest counting confidence method on the test set and compared it with the baseline from our previous work [2]. Furthermore, to assess the influence of each factor on counting confidence, we controlled them separately during training and testing. The results are shown in Table 4, with MSE and $R^2$ (Coefficient of Determination) as evaluation metrics.

Table 4. The comparative results of the pest counting confidence evaluation across different factors

| Factors | MSE | $R^2$ |
|---|---|---|
| PDU | 0.0112 | 0.1135 |
| MDCBB | 0.0107 | 0.1524 |
| AGM | 0.0104 | 0.1737 |
| IC | 0.0094 | 0.2559 |
| IQA | 0.0092 | 0.2722 |
| PN | 0.0090 | 0.2869 |
| Baseline (MDCBB+PN+AGM) [2] | 0.0041 | 0.6740 |
| Ours | **0.0028** | **0.7765** |

As can be seen from Table 4, 1) Among individual factors, pest predicted number, image quality, and image complexity have the highest explanatory power for counting confidence, with $R^2$ values of 0.2869, 0.2722, and 0.2559, respectively. In contrast, the rest of the factors show weaker explanatory power, but still contribute with R² values above 10%.

Therefore, all these factors are worth considering as important indicators for measuring counting confidence. 2) Compared to the baseline, our proposed method shows significant improvements: MSE decreased by 31.7% (from 0.0041 to 0.0028), and R² increased by 15.2% (from 0.6740 to 0.7765).

## 5. CONCLUSIONS AND DISCUSSION

In this paper, we proposed a method for comprehensively evaluating pest counting confidence, based on information related to counting results and external environmental conditions. First, a pest detection network counted pests and extracted counting result-related information. It then applied a hypothesis-driven multi-factor sensitivity analysis to determine the optimal IQA and ICA methods for evaluating image quality and complexity. And an adaptive DBSCAN algorithm was proposed to assess pest distribution uniformity. Additionally, the changes in image clarity induced by stirring during image acquisition are quantified by calculating the average gradient magnitude. Finally, all gathered information from these factors is input into a regression model to predict counting confidence. Experimental results showed each factor significantly impacts counting confidence ($R^2 > 10\%$), and our method outperforms the baseline, reducing MSE by 31.7% and increasing $R^2$ by 15.2%. However, there are some limitations in this paper: 1) The hypotheses and models developed in this paper are based on bionic insects in laboratory settings, rather than real pests in field environments. Real pests are smaller in size, more complex in distribution, and exist under more complex conditions, which lowers detector accuracy. As shown in our previous work [2], the detection model for real aphids achieved an AP@0.5 of 74.8%, compared to 97.1% for bionic insects in this paper. While reduced detector accuracy may affect the explanatory power of the counting confidence model, it does not impact the model's formulation. Further validation with real pests in field conditions will be carried out in future work. 2) Relying on human perception to determine the timings (T2, T3, T4) may lead to misaligned data and affect later analysis. In future work, we will use a robotic arm for stirring and set T2, T3, and T4 at fixed intervals to ensure consistency across all groups of collected samples. 3) The potential impact of detection and segmentation failures of the stirring tool was not considered. In future work, we will test and compare the change in the explanatory power of the counting confidence model with and without removing the stirring tool. Additionally, we are considering excluding the data collected during the stirring phase and focusing only on the data collected after the stirring tool is removed, in order to avoid potential interference caused by the stirring tool.

## ACKNOWLEDGEMENTS

This work was supported by the Engineering and Physical Sciences Research Council [EP/S023917/1], the AgriFoRwArdS CDT, and the BBRO.